\documentclass[12pt,a4paper]{scrartcl}

\DeclareOldFontCommand{\bf}{\normalfont\bfseries}{\mathbf}

\usepackage[utf8]{inputenc}

\usepackage{cite}

\addtokomafont{disposition}{\rmfamily}
\usepackage{newpxtext,newpxmath}
\usepackage{multirow}

\addtokomafont{disposition}{\rmfamily}

\usepackage{nameref}

\usepackage[right]{lineno}

\usepackage{microtype}

\usepackage{changepage}

\usepackage[aboveskip=1pt,labelfont=bf,labelsep=period,singlelinecheck=off]{caption}

\makeatletter
\renewcommand{\@biblabel}[1]{\quad#1.}
\makeatother

\usepackage{color}

\definecolor{Gray}{gray}{.25}

\usepackage{graphicx}

\usepackage{sidecap}

\usepackage{wrapfig}
\usepackage[pscoord]{eso-pic}
\usepackage[fulladjust]{marginnote}
\reversemarginpar

\usepackage{url,lineno,microtype,subcaption}
\usepackage{textcomp}
\usepackage[onehalfspacing]{setspace}
\usepackage[english]{babel}
\usepackage[capitalize, noabbrev]{cleveref}
\usepackage[square,sort,comma,numbers]{natbib}

\begin{document}
\vspace*{0.35in}

\begin{flushleft}
{\Large
\textbf\newline{Dancing Honey bee Robot Elicits Dance-Following and Recruits Foragers}
}
\newline
\\
Tim Landgraf\textsuperscript{1,*},
David Bierbach\textsuperscript{2},
Andreas Kirbach\textsuperscript{3},
Rachel Cusing\textsuperscript{1},
Michael Oertel\textsuperscript{1},
Konstantin Lehmann\textsuperscript{3},
Uwe Greggers\textsuperscript{3},
Randolf Menzel\textsuperscript{3},
Raúl Rojas\textsuperscript{1}
\\
\bigskip
\bf{1} Dahlem Center of Machine Learning and Robotics, Department of Mathematics and Computer Science, Freie Universität Berlin, Berlin, Germany
\\
\bf{2} Department of Biology and Ecology of Fishes, Leibniz-Institute of Freshwater Ecology and Inland Fisheries, Berlin, Germany
\\
\bf{3} Institute for Biology-Neurobiology, Freie Universität Berlin, Berlin, Germany
\\
\bigskip
* tim.landgraf@fu-berlin.de

\end{flushleft}

\section*{Abstract}
The honey bee dance communication system is one of the most popular examples of animal communication. Forager bees communicate the flight vector towards food, water, or resin sources to nestmates by performing a stereotypical motion pattern on the comb surface in the darkness of the hive. Bees that actively follow the circles of the dancer, so called dance-followers, may decode the message and fly according to the indicated vector that refers to the sun compass and their visual odometer. We investigated the dance communication system with a honeybee robot that reproduced the waggle dance pattern for a flight vector chosen by the experimenter. The dancing robot, called RoboBee, generated multiple cues contained in the biological dance pattern and elicited natural dance-following behavior in live bees. By tracking the flight trajectory of departing bees after following the dancing robot via harmonic radar we confirmed that bees used information obtained from the robotic dance to adjust their flight path. This is the first report on successful dance following and subsequent flight performance of bees recruited by a biomimetic robot.

\linenumbers

\section*{Introduction}
One of the most fascinating animal communication systems is the honey bee “dance language” \citep{von_frisch_dance_1967}. Forager bees that found a valuable resource, either food, water, resin, or a suitable nest cavity, might advertise the resource to other nest mates by “dancing” on the comb, or the swarm \citep{lindauer_schwarmbienen_1955, seeley_honeybee_2010, von_frisch_dance_1967}. By actively following the dances, interested foragers can obtain information regarding the resource’s location, profitability and scent. Dances advertising relatively distant resources (> 100 m) exhibit two distinct phases, the “waggle run” in which the bee wags its body laterally while moving forward, and the “return run” in which the dancer circles back to approximately where she started waggling. The waggle portions essentially encode the distance and direction of a resource relative to the hive: The larger the distance of the resource the longer the waggle run in duration and number of wagging movements. The direction of the resource relative to the sun’s azimuth is encoded by the dancer’s body orientation on the vertical comb with respect to gravity \citep{von_frisch_dance_1967}.

But how does a potential recruit decode these properties in the darkness of the hive? Dance-followers track the body movements of the dancer and may touch her body with their antennae \citep{MenzelHoneybeesnavigate2005}. Vibrations of thorax and wings in the waggle run produce air particle oscillations \citep{esch_uber_1961,wenner_sound_1962, michelsen_acoustic_1987,tsujiuchi_dynamic_2007, hasegawa_how_2011}, comb vibrations \citep{autrum_vergleichende_1948, sandeman_transmission_1996, tautz_what_1998, nieh_behaviour-locked_2000}, continuous air flows \citep{michelsen_acoustic_1987,herbert_heran_wahrnehmung_1959, MichelsenSignalsflexibilitydance2003}, and modulated electric fields \citep{greggers_reception_2013,warnke_physikalisch_physiologische_1973} - all of which might be perceived with respective mechanosensors on the cuticula, or in the legs and antennas. Chemical cues, such as environmental odors that cling to the dancer’s body, the taste and scent of the nectar and dance-specific semio-chemicals \citep{breed_recognition_1998-1, gilley_hydrocarbons_2014, thom_scent_2007} characterize the target and might increase foraging motivation or might even help in keeping track of the dancer in the dark hive. Higher body temperature is specific to dancers \citep{FarinaChangesthoracictemperature2001, SadlerHoneybeeforager2011, StabentheinerThermalbehaviorwagtail1995, StabentheinerSweetfoodmeans1991a} and might serve similar functions. Subsequent to following several dance bouts, a follower may exit the hive for a foraging trip. A portion of the recruits reaches the communicated location and joins the collective foraging effort \citep{esch_how_1970, mautz_kommunikationseffekt_1971}. After decades of honeybee dance research, it still remains unknown which of the associated cues play an essential role in attracting, motivating, and instructing the future recruit.  

One way to investigate dancer-follower interactions and study the role of cues emitted by the dancing bee is to substitute the dancer by a biomimetic robot that performs dances in a standardized fashion. This allows the experimenter to pinpoint which of the many cues carry essential information, which stimuli are rather optional or redundant, and which are just by-products of the dance performance. In an experiment by Michelsen et al.\citep{michelsen_how_1992}, bees were found to increase their search for food near a location that was indicated by a mechanical dancer. However, this study did not report any dance-following, nor did the authors videotape the in-hive interactions of robot and dance-followers. Furthermore, the feeding place for the dancer and the test places for the recruits were odor-marked. It remains, thus, unknown whether and how the reported recruits acquired the relevant information from the robot, and whether the recruits observed at the test places were guided by the odor. Furthermore, the authors could not track the flights of the recruited bees leaving some uncertainty regarding how the recruits found the advertised goal. 

In this study, we present the first report on successful dance-following behavior of bees with a robotic bee (hereafter ‘RoboBee’). We give a detailed statistical description of the behavior and compare it to dance-following of natural dances. In addition, we tracked some of the recruits on their consecutive flights via harmonic radar. This enabled us to investigate whether our RoboBee was able to convey spatial information to following bees effectively. In this study we show how bees respond to a dancing robot on three levels. (1) Do bees follow robotic dances at all? (2) How similar is the following behavior to dance-following of natural dances? (3) Can robot-recruited bees extract directional information?

\section*{Materials and Methods}
\subsection*{Study organism and experimental site}
\begin{figure}[h!]
\begin{center}
    \includegraphics[width=10cm]{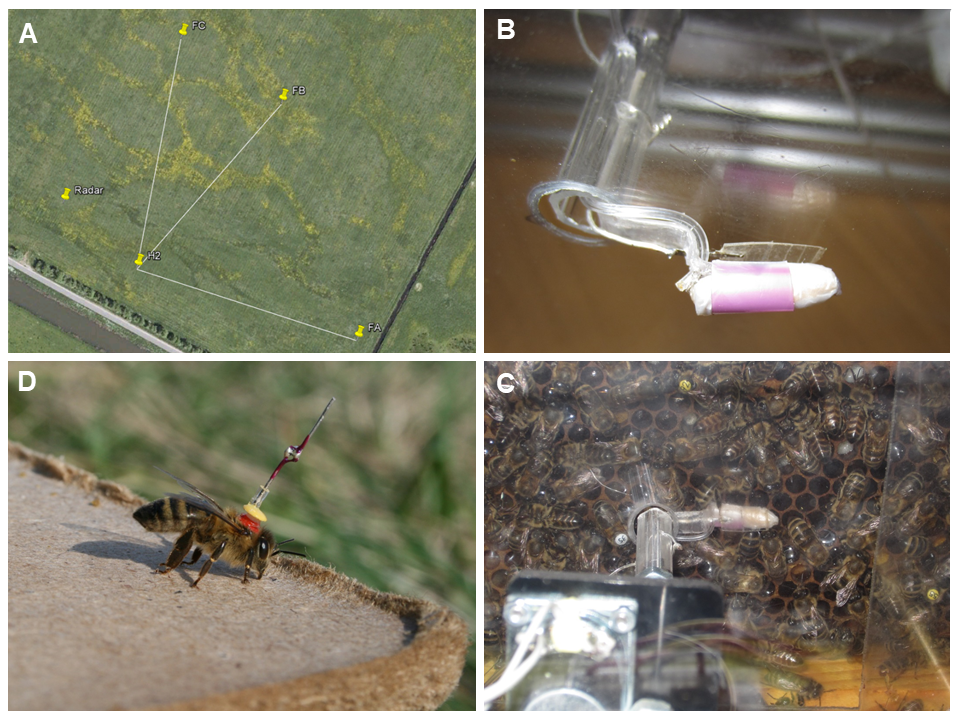}
\end{center}
\caption{The experimental arrangement. (A) The study site, a large pastry close to the village Klein Lüben. (B) A close-up of the bee replica. (C) View from behind RoboBee when into the hive. (D) A honeybee carrying a transponder for tracking with the harmonic radar.}\label{fig:1}
\end{figure}

The experiments reported here were conducted on private grounds leased from a local farmer near the village Klein Lüben. The GPS coordinates of the field site are N52.97555, E11.83677. No further permission was necessary. \cref{fig:1}A depicts the locations of the harmonic radar device, the hive, and the artificial feeders. The radar device and two cabins were placed close to the ridge of the field. One cabin was used for housing the radar console and the person supervising it. The other cabin contained the hive, robot and video recording equipment. Experiments were conducted between end of August and mid-September 2011. The field was mowed in June and, thus, offered only low amounts of natural food sources. 

We used a standard two-frame observation hive with approximately 3000 -- 4000 bees (Apis melifera carnica). We replaced one of the glass panes with a transparent plastic plate and cut a rectangular opening (15 cm x 15 cm) close to the hive entrance, an area often called “dance floor”, since most dances would typically occur there. In this area we filled the space between comb and hive frame with wooden latches to hinder entering bees from changing the comb side quickly. Thus, most natural dances occurred on the side under surveillance.

\subsection*{RoboBee}
\subsubsection*{General setup}
The bee robot is based on a positioning device in three dimensions (planar translation and rotation). It controls the pose of a life-sized honeybee replica (\cref{fig:1}B). The robot stands upright, aligned to the hive. This way, the replica can be moved in parallel to the comb surface (\cref{fig:1}C). The positioning system is based on a plotter (Roland DXY-1300). We replaced its control electronics, cut out most of the plot surface to have a better view on the comb and added a third stepper motor to the pen carriage for rotational motion \citep{landgraf_biomimetic_2010}. The rotation motor carries a plastic rod on whose tip the bee replica is attached. It is made of a soft sponge, wrapped in a small sheet of PE plastic foil. A multi-layered, plastic wing imitation can be vibrated with a loudspeaker attached to the central metal rod. Its vibrations are transduced via a carbon fiber rod. It connects to a metal rocker which is attached to the plastic wing. The wing vibrates radially, producing laminar air flows towards the abdomen. The replica offers tiny drops of sugar water to interested bees, administered manually through a small flexible tube which is routed through the replica with its end emerging at the head. A number of replicas, consisting of body, wing, and tube, were kept in a wire-mesh container several days prior to the experiments to acquire the scent of the hive. To substantially reduce mechanical noise, the 13 Hz waggle motion is produced by the rotation motor only. Therefore, we attached the bee replica eccentrically with respect to the rotation axis. Hence, the angular amplitude and the translational displacement of the body are coupled. Typically, we use 20 mm eccentricity. We introduce the robot into the hive through the window in the hive’s plastic cover plate. To reduce the impact to the hive’s microclimate the robot carries a light-weight transparent plastic sheet to cover the opening. The sheet is uncoupled from the central rod and does not follow rotations, however, it moves with all translations. The entire system is encased in a tiltable aluminum frame, such that it can be aligned to the vertically standing observation hive. Our system thus produces (a) waggle dance motions, (b) wing vibrations and (c) samples of sucrose solution. 

\subsubsection*{Motion model of the waggle dance}
\begin{figure}[h!]
\begin{center}
    \includegraphics[width=10cm]{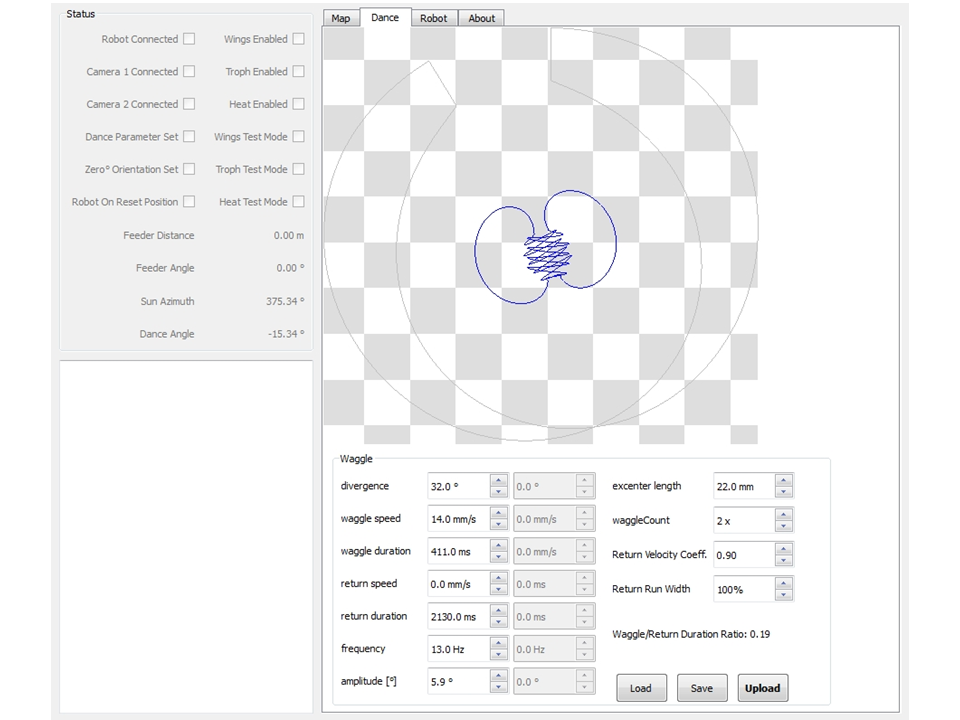}
\end{center}
\caption{Screenshot of graphical user interface of the robot control software. The control software allows altering all dance model parameters and displays a graphical representation of the corresponding trajectory. 
}\label{fig:2}
\end{figure}

We derived a motion model (\cref{fig:2}) from a large database of natural dance trajectories \citep{landgraf_analysis_2011-1}. The model integrates the eccentricity of the bee replica and all other relevant properties of the robot’s hardware. It runs on the robot’s microcontroller and generates the control signals for the three motors to replicate the dance motion. Initial model parameters were obtained through observation of live dancers advertising a feeder location at 230 meters from the hive (data taken from \citep{landgraf_analysis_2011-1}). In order to reduce mechanical noise, we reduced the return run velocity by 40\% (reference: 20 mm/s, robot: 12 mm/s). With lower forward velocity in the return run, the arc described by the robot was smaller than in the reference case. Since the advertised food sources were slightly closer to the hive, the waggle duration was set to 411 ms. The ratio of waggle and return duration encodes food profitability \citep{Hrncirrecruiterexcitementfeatures2011, SeeleyDancingbeestune2000a}, hence, the robot signaled a slightly less profitable source as dances in our reference dataset (reference: 440 ms / 2130 ms = 0.21, experiment: 411 ms / 2130 ms = 0.19). In Landgraf et al. \citep{landgraf_analysis_2011-1} we found the dancer’s body orientation to oscillate in a waggle run with a peak-to-peak amplitude of around 14°. As described above, the robotic waggle was reproduced by vibrating only one motor and affixing the replica 22 mm away from the center of rotation. In order to reproduce a natural peak-to-peak displacement of the replica (3 mm) we reduced the orientation amplitude to 6°.All model parameters are given in Table \ref{tab:1}.

The robot was connected to a PC via USB and a custom control software was used to configure the dance shape and the cues to be emitted by the robot. 

\begin{table}[h!]
\centering
\begin{tabular}{lcc}
 Parameter  & Natural Dance from \citep{landgraf_analysis_2011-1} & RoboBee Dance  \\
 \hline
 \hline
 divergence & 33 ± 19°  & 32° \\
 \hline
 waggle run velocity & 15 ± 5 mm/s  & 11 mm/s \\
 \hline
  waggle duration & 440 ± 160 ms  & 411 ms \\
 \hline
  waggle frequency & 12.67 & 13 Hz \\
 \hline
 waggle amplitude & 13.89° | 2.64 mm  & 5.9° | 3 mm \\
 \hline
  return run velocity & 20 ± 4 mm/s & 12 mm/s \\
 \hline
  return duration &2130 ± 470 ms  & 2130 ms \\
 \hline
  waggle/return ratio (duration) & 0.21  & 0.19 \\
 \hline
\end{tabular}
\caption{Dance parameters used for the RoboBee during the experiments in comparison to parameters obtained from the observation of natural dances. Properties given without standard deviation were not following a normal distribution.}\label{tab:1}
\end{table}

\subsection*{Pre-training of foragers}
We pre-trained bees to artificial feeding sites. Using the robot to advertise a previously visited food location might reduce the complexity of the decoding task for the dance-followers and might increase the bee’s motivation to fly out and visit this feeder. Starting with the 26th of August, we provided 2 M sucrose solution alternatingly at two artificial feeders (FA and FB, 200 m away from hive, see \cref{fig:1}A). On the first days of training, bees that were foraging at one feeder were caught and carried to the other feeder. All bees that were allowed to drink sucrose were number-tagged on the thorax with a colored number plate displaying a unique two-digit ID. 

A continuous protocol was kept of all bees that were rewarded at FA or FB, respectively. Starting with the 30th of August, newcomers at the feeders were number-tagged and paint-marked but were not brought over to the other feeder anymore. Hence, among the tagged bees, there were some that experienced both feeders and some that only visited FA or FB. Feeders were removed immediately after the feeding hours. In total, N=193 different bees were tagged over the course of the experiment. Prior to testing the dances of RoboBee, the feeding stations and other visual cues were removed from the field.

\subsection*{The experiment}
RoboBee was configured to perform dances to either one of the known feeders, or in some instances a virtual feeder, “FC”, which was 30 degrees from FB, and 90° from FA (see \cref{fig:1}A). Even without recruitment dances, bees revisit the feeders periodically. Bees having acquired experience with both feeding locations, however, show a strong propensity to first visit the feeder that was previously rewarded. Hence, we configured RoboBee to advertise the feeder that was not rewarded before (either FA or FB) or in some instances FC. During the test, no sugar solution was provided at any of the field locations. To change the advertised location of the feeder, we changed the parameter “dance angle” according to the given field location, date and time. The dance angle was updated every 5 minutes to account for the shift of the sun’s position. All other parameters remained fixed. The robot was operated in sessions of variable duration, interrupted for up to 30 minutes for several reasons, depending on the level of colony activity and acceptance of the robot. For example, in the morning hours (9:00 am – 11:00 am) the foraging activity of the colony was often low due to environmental conditions (low temperature, high humidity). As an effect, the general interest towards the robotic dancing was low and therefor paused for up to 30 minutes. The robot was “parked” in a corner of the dance floor. In some occasions, bees displayed aversive behaviors towards the robot during dancing. The dances were paused during those events as well. Under optimal conditions the dances were continuously performed for 5 - 10 minutes, followed by an equally long break. When bees showed high interest in the RoboBee, the dance sessions were not interrupted. 

Four persons conducted the experiments. The robot operator controlled the robotic dances by means of a keypad. The dances could be interrupted, resumed, and shifted along the x and y axes. One observer sat close to the hive and reported the IDs of bees that showed lively interest in the robot, i.e. running after the robot without displaying aversive behaviors such as climbing and holding onto the robot. When one of those bees walked towards the exit of the hive, a respective signal was given to a third person, the bee handler, outside of the cabin who observed the hive entrance, took note of the reported IDs and waited for the announcement of a leaving bee. He then caught the bee, fixed a transponder on its thorax (see \cref{fig:1}D) and released it immediately. Afterwards, he reported the ID to the supervisor in the radar cabin where the time, identity, and the radar signals of the respective flight were recorded.

\subsection*{Video recording and analysis of the dance-following behavior}
All in-hive performances of RoboBee were audio and video recorded at 50 fps using diffuse daylight. The camera (Basler A602f) was set up to observe the entire dance floor. Due to low sensor resolution (640 x 480 pixels), the identity of tagged bees could not be extracted from video reliably. We therefore commented on audio when relevant bees were close to the robot. A reference video dataset of natural waggle dances and dance-following was analyzed analogously. The reference dataset exhibits higher spatial and temporal resolution. The robot videos have a resolution of 3.4 px per mm at 50Hz, the natural dances have a resolution of 7.8 px per mm at 100 Hz. Furthermore, the camera’s viewing angle was directed perpendicularly to the comb surface. The video dataset can be found online \footnote{\url{https://www.youtube.com/watch?v=iCbGrvkKzxo&list=PLs7Vp-pCDX7zTZxsrwsnKX-oj7gYC7pCW}}.

In the video analysis, we searched for bees showing dance-following behavior. We defined dance-following similar to \citep{bozic_attendants_1991}. Besides a high motivation to stay in close proximity to the dancer, we looked specifically for bees that were eager to touch the wagging abdomen of the dancer and follow its return runs, participating in the turns such that the follower herself describes alternating rotations of almost 360 °. 

For each bee that displayed following behavior, the number of continuously followed waggle runs was scored from video manually. Due to high numbers, short following behavior (less than 3 waggle runs) was disregarded. We registered a following behavior, if the animal followed three waggle runs or more. The shortest sequence, therefore, was waggle – return – waggle – return – waggle. If the animal missed one waggle run (defined as the distance of her head to the dancer’s body exceeding half a body length for more than one second), the sequence was regarded as interrupted. If afterwards the following behavior was resumed, it was registered as a new sequence (only if, again, more than two waggle runs were followed). 

\begin{figure}[h!]
\begin{center}
    \includegraphics[width=15cm]{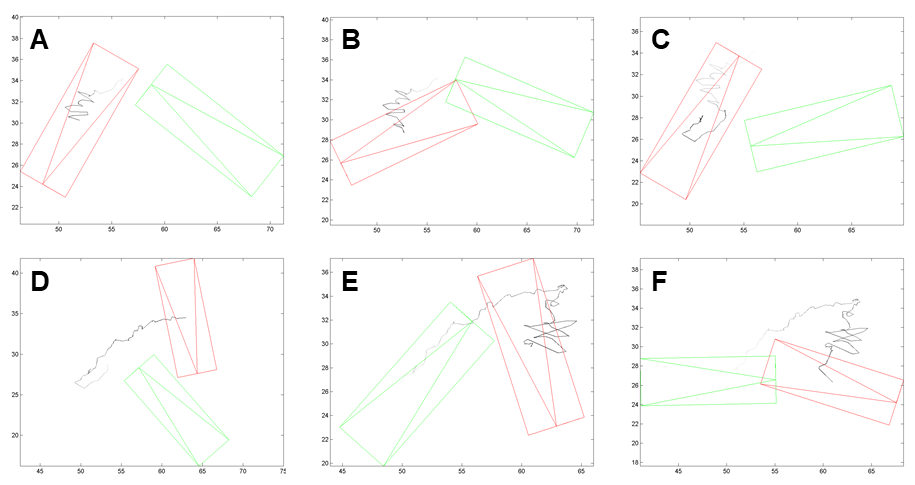}
\end{center}
\caption{A typical dance-following trajectory. At the time of the waggle run, the follower bee often stands in a perpendicular configuration (A) The dancer, denoted by the red box, turns clockwise into her return run, the follower remains at the abdomen (B) and changes from the dancer’s left to her right side (C) On her way to the next waggle the dancer is followed almost perpendicularly. Often, the dancer turns a little faster and the follower ends up facing the dancer for a short period of time (D) Quickly, the follower continues her turn to reach the perpendicular configuration in the waggle period (E) The dancer turns into the return run circling counter-clockwise this time. The follower analogously switches sides (F) and continues as described above.
}\label{fig:3}
\end{figure}

All video sequences containing following behavior as defined above were processed with a custom tracking program \citep{landgraf_tracking_2007}. For each frame during a waggle run, and for every 10th frame in return runs, a rotatable bounding box was set to approximate the position and orientation of dancer and followers (\cref{fig:3}). We tracked each animal until they either left the borders of the video frame or stopped showing following behavior. 

Since the camera’s viewing angle was not perpendicular to the comb surface, the resulting videos exhibited perspective distortion. To rectify the motion sequences, we manually determined the image positions of the four corners of the aperture through which the robot was inserted for each video file. We then calculated the homography matrix H which can be used for mapping image coordinates $x_i$ to real world coordinates $Hx_i$ = $x_r.H$ was determined with third party software in Matlab \citep{kovesi_matlab_2000}. In a second preprocessing step, we interpolated the trajectory data to “fill” gaps in the return runs. The interpolation restored the original sampling rate of 50 Hz in the return runs. In the last step, the frame indices of start and end times of the robot’s waggle runs were manually extracted. This information was extracted automatically in the reference dataset (see \citep{landgraf_analysis_2011-1} for details). These indices were used to cut trajectories into motion sequences starting with a waggle run. In the following analyses we processed two types of motion sequences:  trajectories that contain one waggle and the consecutive return run (called “half period”) or two waggle and return runs (“full period”). 

Due to individual variability, the duration (i.e. the number of elements in the vector) of the resulting motion sequences varied significantly. Hence, we resampled the data to a unitary duration. Note that, due to the resampling, the time axis becomes unitless, “0” corresponds to the beginning of the waggle, “1” denotes the end of the return run.

Each motion sequence was transformed to a vector with the same number of elements. For later comparison, video recordings of natural dances and following were tracked and processed analogously (data from \citep{landgraf_analysis_2011-1}). 
We extracted several features from the motion sequences to compare the behavior of followers of RoboBee with dance-followers of natural dances. First, we computed the Euclidian distance of the follower’s head to the body axis of the dancer. Second, we calculated the cosine of the mutual angle between the two longitudinal body axes. A head-to-head configuration results in negative values, a perpendicular pose yields values close to zero and positive values represent an almost parallel alignment of the two bodies. Note that a negative cosine could also mean the dancer and follower are facing away from each other in an abdomen-to-abdomen configuration, reflected by a large head-to-body distance. Third, we translated the trajectories into an ego-centric coordinate frame and computed forward, sideward and turning velocities through a full waggle period. These ego-centric motion velocities were then averaged over all sequences and integrated to compute the mean position of the follower throughout a full waggle cycle. 

\subsection*{Tracking the flight of recruits}
The working principle of the harmonic radar system has been described and improved over the last decades \citep{Riley_1996}. Dance-followers of robotic dances were identified, caught and fitted a radar transponder before they could begin their foraging trip. Due to the lack of an appropriate interface, the output of the radar console was captured visually with screen grabbing software which saved the screen image as bitmap files once every second. The trajectories were then obtained with custom tracking software. The corresponding output was plotted with R scripts and edited with Adobe Illustrator CS5.

\section*{Results}
\subsection*{Bees followed RoboBee’s dances}
We detected following behavior in 8 of 13 days with a maximum of 29 dance-following instances on 31st of August. Only a portion of these animals were marked, hence, it is unknown to how many different individuals this corresponds. Dance-following was observed between 11:48 and 16:58. In total, we observed 80 dance-following instances over the entire period. Two bees followed simultaneously in two separate instances. Only six of the dance-followers were individually tagged (see \cref{tab:2}), the rest were unmarked bees. RoboBee’s dances were followed significantly longer than natural dances (number of followed waggle runs per dance, RoboBee: 7.19 $\pm$ 3.73; live dancer: 5.45 $\pm$ 2.77; MW-U-Test; U=1211, P=0.004, N1=80; N2=44). Please note that we included only dances in our analysis in which 3 or more waggle runs were followed (see methods). 

\begin{table}[h!]
\centering
\begin{tabular}{lccc}
 \multirow{ 2}{*}{} ID  & Total number of  & Number of days followed  & Radar flights \\
& waggle runs followed & (days in between without following) &recorded  \\
 \hline
 \hline
 \#45 & 51  & 2(0) & No \\
 \hline
  ng71 & 47 ± 19°  & 2(1) & Yes \\
 \hline
  ng62 & 31 ± 19°  & 1 & Yes\\
 \hline
  ny18 & 29 ± 19°  & 1 & Yes\\
 \hline
  \#70 & 18 ± 19°  & 1 & No\\
 \hline
  ny47 & 8 ± 19°  & 1 & Yes\\
 \hline

\end{tabular}
\caption{Six of dance-followers that were video recorded were marked. For those bees we could verify whether they repeatedly followed RoboBee's dances. Most of the waggle runs were followed during one day. Two animals were observed to remain interested in the robot for longer.}\label{tab:2}
\end{table}

\subsection*{Comparing following behavior}
\begin{figure}[h!]
\begin{center}
    \includegraphics[width=15cm]{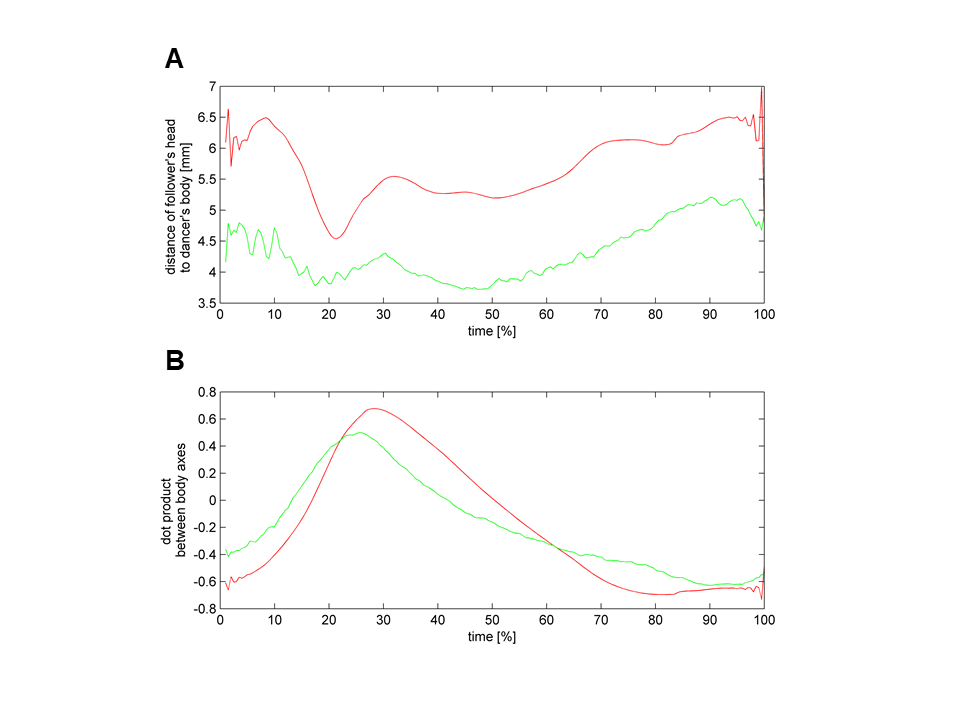}
\end{center}
\caption{(A) Distance and relative angle of follower to dancer over waggle and return run. The average distance of the follower’s head to the dancer’s central body axis is depicted for robotic and natural dances in red and green, respectively. (B) The average dot product of body direction vectors of dancer and follower. Both figures refer to a unit period of time. Time t = 0 corresponds to the start of the waggle, t = 1 denotes the start of the ensuing waggle. The red curves show the data extracted from robotic dances; the green data displays the reference data extracted from natural dance recordings.
}\label{fig:4}
\end{figure}

We collected 155 full periods in which bees followed RoboBee and compared them to 88 full periods in which bees followed live dancers. The head-to-body distance was on average 1 mm larger for RoboBee sequences (\cref{fig:4}A), however, the shape of the time series exhibits similar features. Throughout the waggle phase (t = 0\% – 10\%), the head to body distance remains approximately constant. When the dancer turned into the return run (t = 10\% – 30\%) the head to body distance dropped to a low value right when the follower switched sides (t = 20\%). Throughout the entire return run (t = 10\% – 100\%) the head-to-body distance stayed smaller than a body length (means for robotic vs. natural dances: \textasciitilde 5 mm and 4 mm, respectively), but tended to increase over time in both datasets. 

Both follower groups show similarities in the course of the mutual body angle (see \cref{fig:4}B). The cosine of this angle starts with negative values reflecting a head-to-head configuration. The cosines then cross 0 right after the waggle run (t = 10\%, corresponds to a perpendicular configuration; Figure 4b), and increase to values close to 1 (t= 20-30\%, when the followers switch sides and look into the same direction) and come back to values close to the starting point (t=60-100\%), which means the followers turn with the dancer, but may simply be too slow to stay in a perpendicular configuration throughout the return run. 

\begin{figure}[h!]
\begin{center}
    \includegraphics[width=15cm]{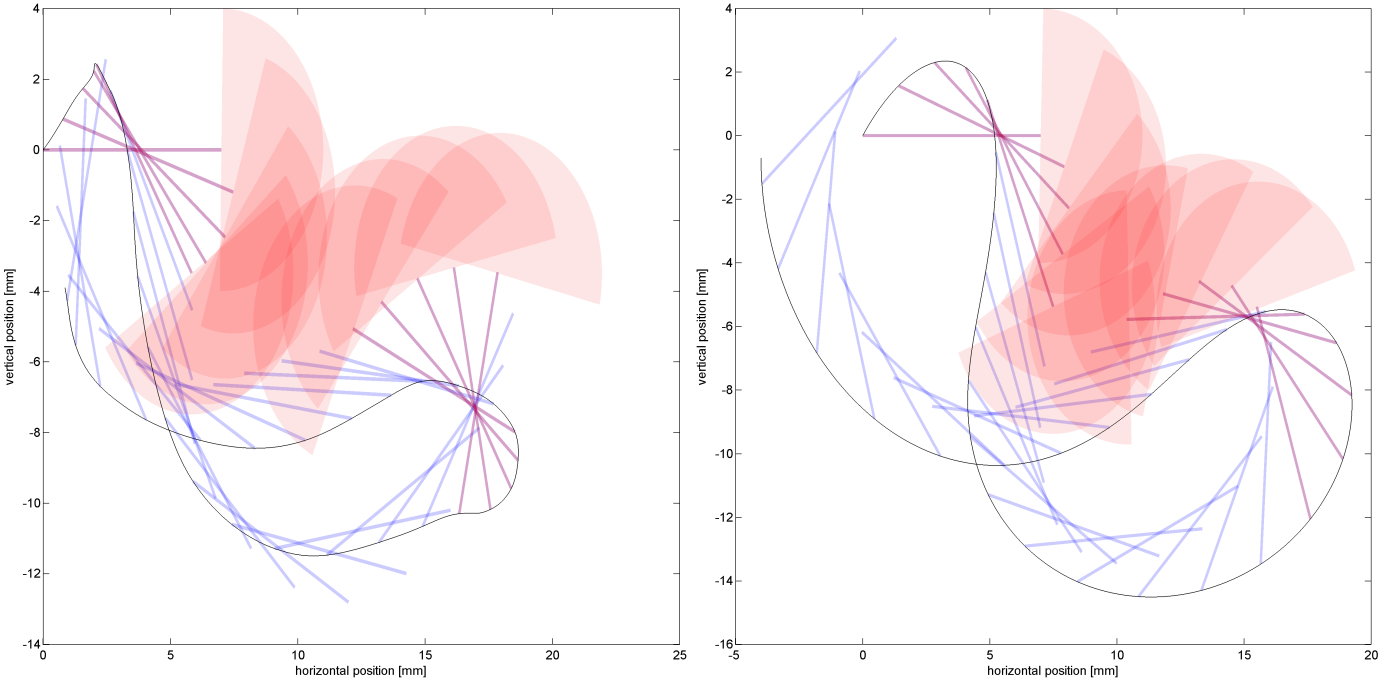}
\end{center}
\caption{Average dance-following trajectory. A virtual dance-follower at position (0,0) with orientation 0° was moved by applying forward, sideward and turning velocities for each time step. The resulting body center positions over the course of a full waggle period (black solid lines) for the dance-followers of RoboBee (left) and natural dances (right) exhibit similar features. In both cases, the dance-follower describe a path resembling the figure 8. The red semi-circles denote the region that dance-followers may be able to touch with their antennae during the waggle portion of the dance. The blue and red lines represent the body orientation of the dance follower. Although not explicitly modeled, the dancers waggling motion points towards the lower right of the figure, and is located where the antennal sensory regions overlap.}\label{fig:5}
\end{figure}

We then analysed the dance-following motion and computed the forward, sideward, and angular velocities of a follower. This data has been extracted from 42 and 70 following runs (including each two waggle runs) from robotic and natural dances, respectively. Each sequence consists of body positions and orientation angles over time for a dancer’s full waggle period (the sequence waggle – return – waggle –return). For the sake of consistency, the first waggle run is followed by a clockwise return run for all sequences. All were resampled to an equal number of pose samples (500 time points in the present study, corresponding to a sampling frequency of approximately 100 Hz). We computed the mean over all 42 robotic and 70 reference sequences. We then integrated these motion velocities to the average dance following path by cumulatively adding consecutive motion vectors to a starting position at (0,0) and 0° body orientation. Both trajectories exhibit a similar pendulum-like motion over similar spatial scales (for details see \cref{fig:5}). 

\subsection*{Evaluation of information transfer}
\begin{figure}[h!]
\begin{center}
    \includegraphics[width=15cm]{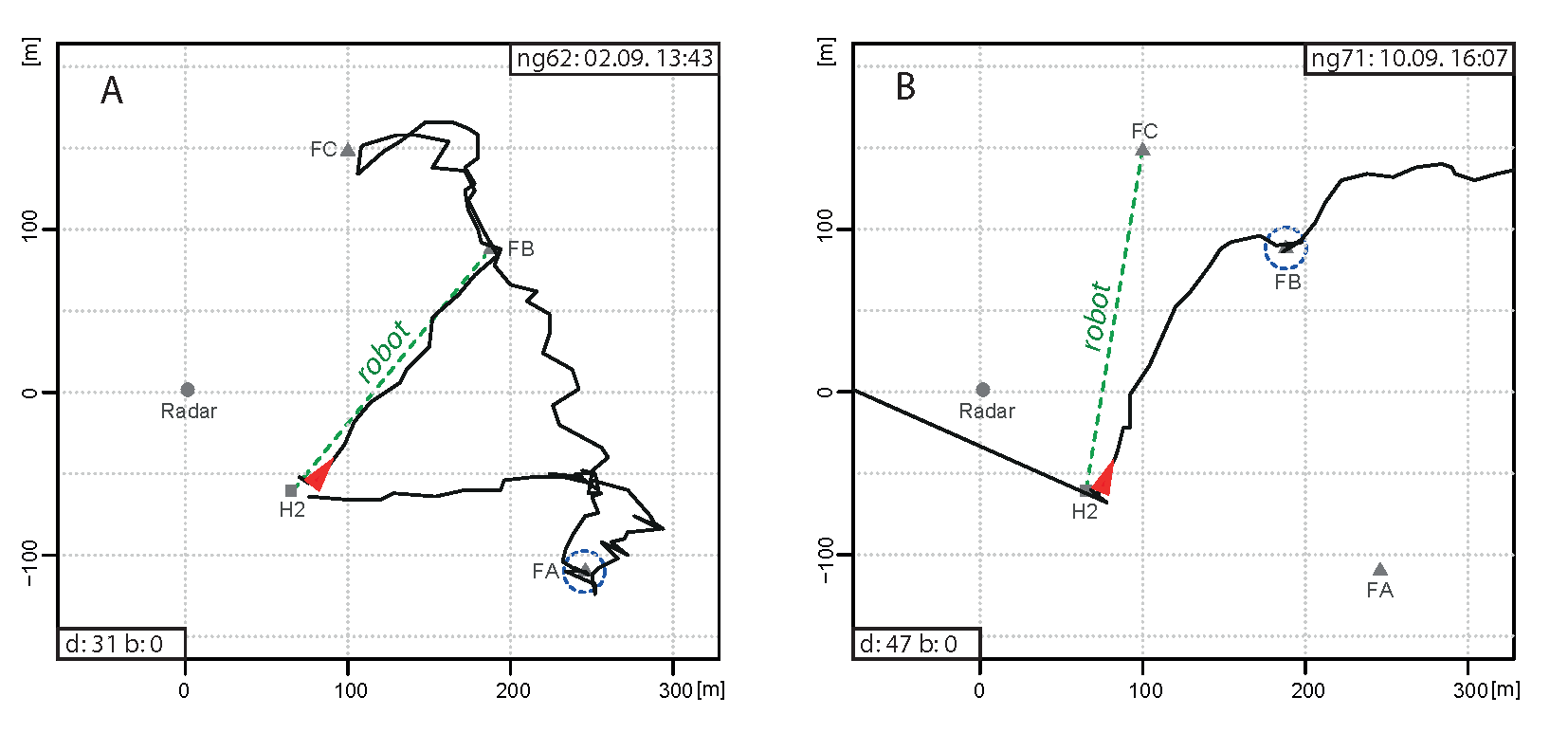}
\end{center}
\caption{Flight paths of bees that followed many robotic waggle runs.  Black lines depict flight trajectories, starting at the red arrow head, which points in the direction of the outbound flight. Feeders marked with a dotted circle have been reinforced the previous training day. The direction which was indicated by the robotic dance is depicted with a green dashed line. In each subfigure the ID of the individual and time of flight are given in the upper right corner. The number of waggle runs followed right before the flight (d) and the accumulated number of waggle runs followed all days before (b) are given in the box to the lower right.}\label{fig:6}
\end{figure}

Only 6 individually tagged bees followed RoboBee’s dances inside the hive, four of which could be radar tracked on their consecutive foraging flight (see \cref{tab:2}). Two of those four bees (ng62 and ng71 shown in \cref{fig:6}) were following dances extensively that pointed to feeding sites they did not know at that point in time. Ng62 had experienced FA only and followed 31 waggle runs pointing to FB. The radar tracks show that she first flew to the unknown location (FB), turned away from FB, flew a 180° loop of about 130 m distance and finally visited FA, the location she had been foraging on the day before. After a short search, she returned to the hive where she was caught and released after transponder removal. Conversely, bee ng71 was rewarded at FB the day before the test. She followed 47 waggle runs pointing to a virtual feeder (FC). On the first 100 m of her outbound trip she flew approximately towards the middle point between FC and FB. She then seemed to converge to FB. Then, the bee started a long search trip outside the observable boundaries and came back after several hours. 

We were able to track more than one flight for two additional individuals. Bee ny18 had records for feeding from both sites, FA and FB. She visited feeder B on the 26th, 30th, 31st of August and FA once on the 31st of August. On that day, both feeders were open; FA from 10:00 am to 12:00 am and FB in the evening from 5:00 pm to 6:30 pm. Ny18 followed the whole day on September 3rd. She first showed light interest (sampling only a few waggles, without motivated following behavior). The radar tracks registered at 2:00 pm show her direct flight to FB, the robot pointing to FA. Prior to her second flight at 4:29 pm she followed 14 waggle runs with high motivation. This time she visited FA. At 4:55 pm, she showed a high motivation to decode the dance. The robot was set to indicate food at feeder B. The bee followed 29 waggle runs (twice as much as before) and flew to FB, then south to F A and then home again. She did not land at the hive (thus wasn’t caught) and flew out again, describing a broad loop over FB, returning home eventually. 

\begin{figure}[h!]
\begin{center}
    \includegraphics[width=17cm]{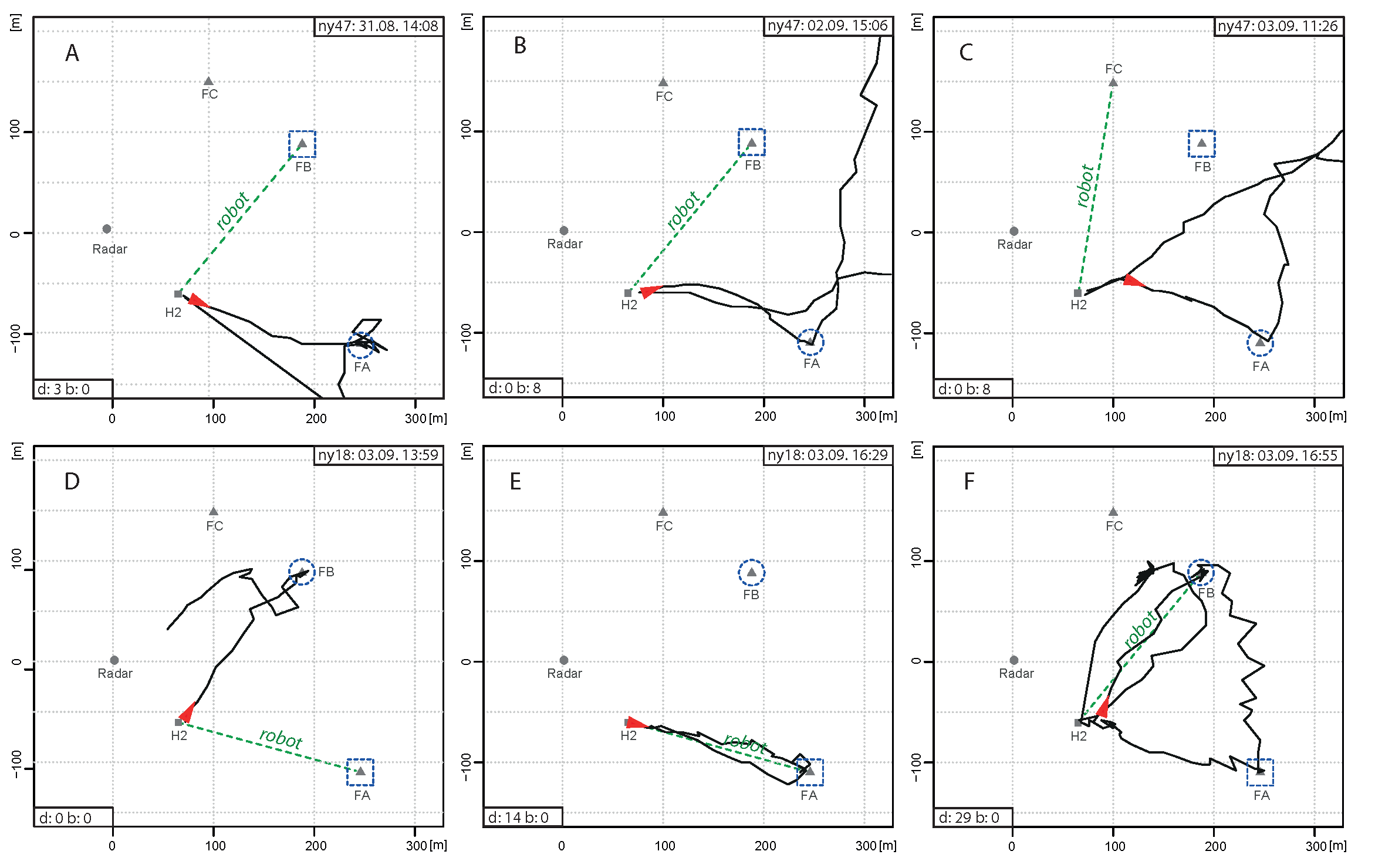}
\end{center}
\caption{Radar tracks of flights of bees with various levels of experience with robotic dances. The first row depicts the flight paths of bee ny47 who, although having accumulated a number of sampled waggle runs, never displayed highly motivated following behavior. Her outward vector flights point to FA, irrespective of the robotic dance that pointed to FB or FC. Bee ny18 (second row) was traced as control bee before having followed waggle runs. Throughout the day, she became very interested in the robot and showed a high number of motivated following runs. She subsequently visited the location indicated in the dance, ignoring her previous experience. Flight paths are depicted as a black line, starting at the red arrow head. Feeders marked with a dotted rectangle have been reinforced at least once in preceding days. Circles mark last rewarded feeders. The green dashed line denotes the direction communicated by RoboBee. In each subfigure, the ID of the individual and time of flight are given in the upper right corner. The number of waggle runs followed right before the flight (d) and the accumulated number of waggle runs followed all days before (b) are given in the box to the lower right. 
}\label{fig:7}
\end{figure}

A bee that showed less interest in the robot was ny47. She showed up near the robot regularly but following could not be elicited before 11:48 am. She followed three waggle runs and did not continue. Although she occasionally came near the robot and had antennal contact to the waggling robot (several times in single waggle runs) she did not run out before 2:29 pm. The radar track shows her visiting the previously visited location FA, the robot indicating FB. Later the same day she followed 5 more waggle runs but could not be traced again. Two days later, she could be caught again having followed no waggle. Radar traces show her visit to FA again. The same happened the next day: no waggles followed, she visited the previously rewarded site. \cref{fig:7} shows the flight tracks of ny47 and ny18.  

\section*{Discussion}
We showed that RoboBee's dances elicited similar behavioral responses as observed in natural dances. Not only do bees follow robotic dances, they follow even for longer periods than with live dancers. The trajectory dance-followers describe in robotic dances is very similar to the one performed in natural dances and may be sustained over dozens of waggle runs. After having followed a robotic dance, the behavior of those recruits appears as expected. Most animals either exit the hive directly, or take up a honey ration shortly before leaving. We were able to track the flights of some of these recruits and all flight paths indicate that directional information was indeed transmitted by the robot. This is the first report of honey bees actively decoding robotic imitations of waggle dances and, although anecdotal in numbers, first evidence that the information encoded in the dance could actually be transmitted to live bees. 

However, in order to recruit and track more recruits’ flights, the honey bee robot has to be improved. Natural dancers are often followed by many bees at the same time, competing for a favourable position relative to the dancer. In only two of the 80 dance-following instances described above, we observed two dance-followers simultaneously decoding robotic dances. This might explain why the average number of continuously followed waggle runs is lower in natural dances. 

The robot attracted dance-followers in only 8 out of 13 days. It still remains an open research question how to improve the robot to attract more followers. However, those individuals that showed dance-following expressed a high motivation to decode the robotic dance. These individuals did not show any less interest in the robotic dance, not did they seem to have difficulties in tracking the robot though waggle and return phases. Their motion dynamics and the integrated average motion path resemble closely their natural dance-following reference. The low number of dance-followers, hence, may indicate a low attractiveness of the robotic dance. Several factors may affect how attractive an animal perceives a dance. Environmental factors, such as weather or the general availability of food, may modulate a forager’s motivation to forage and also to follow dances \citep{al_toufailia_honey_2013,dreller_regulation_1999,seeley_social_1989,seeley_wisdom_1995}. Adding to that, personal experience has been shown to modulate the individual behavior of dance followers and recruits \citep{gruter_honeybee_2009,balbuena_honeybee_2012,gruter_informational_2008}. Robotic dances may furthermore not have fully reproduced all relevant cues, such as thermal cues \citep{tautz_honeybee_1996}, or may have produces aversive cues such as odors emanating from the materials used.  

Nonetheless, we were able to track some of RoboBee’s recruits on their ensuing foraging flights by harmonic radar. Although the number of those flight records is low (4 individuals, 8 tracked flights), we tracked bees that were naïve to RoboBee’s advertised locations. While one of those bees was observed to fly to the indicated location after having followed 31 waggle runs, the other bee, having followed 47 waggle runs, visited a previously experienced location. Interestingly, her flight started into the direction indicated by RoboBee, but soon converged to the known location only 30 degrees away. Such a behavior is also known from bees that followed natural dances and Menzel et al. \citep{menzel_common_2011} report flights of bees that averaged private and social information to a balanced flight direction. As a control, bees that did not follow RoboBee (ny47 on all three radar-tracked flights, ny18 on her first radar-tracked flight) flew to the food source they had experienced beforehand. 

In conclusion, our results indicate that live bees were able to extract directional information from RoboBee’s dances. This is the first report of extensive dance-following and the first dataset of flight trajectories of bees that were instructed by a dancing robot. In comparison to their natural counterparts, robotic dances may be perceived less attractive and, given the complex experimental procedure, the robot has to be improved to be able to record a significantly larger number of radar tracks. The right choice of materials, as well as a better chemical camouflage will be essential in future prototypes. The next version of RoboBee will be integrated into the comb. It will be part of the hive structure and thus chemically indiscriminable from the hive. Although the systematic improvement of RoboBee might require extensive resources, we believe that tools such as RoboBee will be very helpful in understanding the fascinating intricacies of honeybee communication.

\section*{Acknowledgments}
We are indebted to Hai Nguyen and Daniel Rhiel for their valuable contributions in the field and Franziska Boenisch for her help in preparing this manuscript.

\section*{Author contribution}
Conceived and conducted experiments: TL, AK, RC, KL, UG, RM, R.
Construction of robot: TL, MO, AK.
Data analysis and statistics: TL, DB, AK.
Writing manuscript: TL, DB, RM.
Supervision: RM, RR, TL.

\nolinenumbers

\bibliography{library}

\bibliographystyle{plainnat}

\end{document}